\documentclass[sigconf]{acmart}

\usepackage{xcolor}
\usepackage{url}
\usepackage{graphicx}
\usepackage{subcaption}
\usepackage{multirow}
\usepackage{xcolor,colortbl}
\usepackage{cleveref}
\usepackage{xcolor,soul}
\usepackage{enumitem}

\AtBeginDocument{%
  \providecommand\BibTeX{{%
    \normalfont B\kern-0.5em{\scshape i\kern-0.25em b}\kern-0.8em\TeX}}}

\copyrightyear{2024}
\acmYear{2024}
\setcopyright{rightsretained}
\acmConference[TAS '24]{Second International Symposium on Trustworthy Autonomous Systems}{September 16--18, 2024}{Austin, TX, USA}
\acmBooktitle{Second International Symposium on Trustworthy Autonomous Systems (TAS '24), September 16--18, 2024, Austin, TX, USA}
\acmPrice{}
\acmDOI{10.1145/3686038.3686067}
\acmISBN{979-8-4007-0989-0/24/09}

\begin{document}

\title{Mapping Safe Zones for Co-located Human-UAV Interaction}

\author{Ayodeji O. Abioye}
\email{a.o.abioye@soton.ac.uk}
\orcid{0000-0003-4637-3278}
\affiliation{%
  \institution{University of Southampton}
  \city{Southampton}
  \country{UK}
}

\author{Lisa Bidgood}
\email{lb11g21@soton.ac.uk}
\orcid{0009-0001-0202-778X}
\affiliation{%
  \institution{University of Southampton}
  \city{Southampton}
  \country{UK}
}

\author{Sarvapali D. Ramchurn}
\email{sdr1@soton.ac.uk}
\orcid{0000-0001-9686-4302}
\affiliation{%
  \institution{University of Southampton}
  \city{Southampton}
  \country{UK}
}

\author{Mohammad D. Soorati}
\email{m.soorati@soton.ac.uk}
\orcid{0000-0001-6954-1284}
\affiliation{%
  \institution{University of Southampton}
  \city{Southampton}
  \country{UK}
}

\renewcommand{\shortauthors}{Abioye et al.}

\begin{abstract}
Recent advances in robotics bring us closer to the reality of living, co-habiting, and sharing personal spaces with robots. However, it is not clear how close a co-located robot can be to a human in a shared environment without making the human uncomfortable or anxious. This research aims to map safe and comfortable zones for co-located aerial robots. The objective is to identify the distances at which a drone causes discomfort to a co-located human and to create a map showing no-fly, moderate-fly, and safe-fly zones. We recruited a total of 18 participants and conducted two indoor laboratory experiments, one with a single drone and the other set with two drones. Our results show that multiple drones cause more discomfort when close to a co-located human than a single drone. We observed that distances below 200 cm caused discomfort, the moderate fly zone was 200 - 300 cm, and the safe-fly zone was any distance greater than 300 cm in single drone experiments. The safe zones were pushed further away by 100 cm for the multiple drone experiments. In this paper, we present the preliminary findings on safe-fly zones for multiple drones. Further work would investigate the impact of a higher number of aerial robots, the speed of approach, direction of travel, and noise level on co-located humans, and autonomously develop 3D models of trust zones and safe zones for co-located aerial swarms.
\end{abstract}





\maketitle

\section{Introduction}
Recent advances in hardware technology, computing capability, and algorithms powering robotics could make ubiquitous robots in our everyday lives a reality in the not-so-distant future. The application of robotics has gone beyond industrial or outdoor environments and can now be found in homes, offices, hospitals, malls, streets, etc. There are growing demands for uses within civilian spaces for applications such as wildlife conservation, agriculture, entertainment, education, care, warehouse management, parcel delivery, facility inspection, construction monitoring, firefighting, and disaster response. They can complement and work side-by-side with humans, collaborating as any other team member. This means robots will be increasingly co-located in the same environment as humans, sharing the same workspace or living space. Therefore, there is a need to define safe zones for robots to operate without disrupting human activities or making them uncomfortable or anxious.

This paper focuses on a particular case of swarm robotics involving multiple drones capable of being localised in three-dimensional spaces around the human user. In this work, we build on existing studies in human-human proxemics \cite{Bretin2023,Hall1990}, human-drone interaction \cite{Cauchard2015,chang_spiders_2017,Abioye2022} and human-swarm interaction \cite{parnell_trustworthy_2023,Soorati2024} to identify safe zones for swarm operation and human-swarm interaction using aerial swarms. The unmanned aerial vehicle (UAV) may need to be close enough to the human to capture speech or gesture with its onboard microphone and camera for a natural human-swarm interaction. Some of the factors that can affect how close an aerial swarm can be to people include UAV size, speed of travel, direction of travel, number of co-located persons, age of co-located person (child or adult), proximity (distance), anxiety, safety, sound (noise), altitude, wind, field of view, frequency, reliability, and level of autonomy. How close is too close? When should a swarm hold it right there? To address this, we conducted a user study (N = 18) using single and multiple 450 mm diameter quadrotor UAVs to map safe and comfortable zones for co-located aerial swarms in natural human-swarm interaction. Our method involves a multiple-path UAV trajectory in which the drone approaches or moves away from the participants at different angles.

This research aims to define safe operating zones for aerial swarms being deployed in human-occupied environments such as homes, offices, schools, hospitals, warehouses, etc. where there is a limited amount of shareable space. By maintaining a minimum separation distance from the human, the swarm could operate safely and cause minimal discomfort.

\section{Related Works}
Proxemics is the study of how humans use space. It is derived from the term `proximity'. It can be classified based on functional or defensive usage \cite{Bretin2023}. Based on functional usage, \cite{Hall1990} stated four distinct distances: intimate (0 - 46 cm), personal (46 - 122 cm), social (122 - 366 cm), and public (365 - 762 cm). The buffer zone and peripersonal space (reaching space around the body) concepts are examples based on defensive usage \cite{Bretin2023}. However, this term has gained popularity beyond human-to-human spacing and can be used in describing human-robot spacing. Mumm and Mutlu \cite{Mumm2011} investigated human-robot proxemics in which participants approached a stationary human-like Wakamaru robot in a laboratory experiment. They found that participants preferred to maintain a distance between 100 cm and 105 cm. They also found that this distance was higher when the participants were in front of the robot, suggesting the participants were uncomfortable and hence needed a higher buffer zone to avoid a sudden robot approach. Duncan and Murphy \cite{Duncan2012} modelled a comfortable distance by combining the research of 19 previous studies. They found that the larger the angle and the slower the speed of approach, the more comfortable people were. They also investigated the effect of the UAV approach at two different human heights (2.13 m and 1.52 m) and found no significant difference in comfort levels \cite{Duncan2013}. In this research, we focus on defining the human spatial boundaries with aerial swarms to ensure safe and comfortable operations that are not disruptive. 

Acharya et al. \cite{Acharya2017} investigated comfortable human-robot distances by conducting a user study with 16 participants to compare a small UAV (AscTec Hummingbird 540 mm diameter quadcopter) with a ground robot (Double Telepresence Robot). In their research, the UAV approached at a fixed speed of 20 cm/s and the participants were required to say `stop' when an approaching robot came too close. They found that the comfortable distances were 36.5 cm for the ground robot and 65.5 cm for the small UAV. Their post-study survey showed that higher UAV noise and wind decrease comfort. Wojciechowska et al. \cite{wojciechowska_collocated_2019} also found that participants preferred a UAV within personal space (50 to 121 cm) over intimate (< 50 cm) and social distances (> 121 cm). They used a Parrot AR Drone 2.0 with dimensions 580 x 580 x 130 mm. Participants observed drone movements, completed surveys after each trial, and underwent post-study interviews. The Self-Assessment Manikin test assessed participants’ emotional states before, during, and after the study. The path followed by the autonomous drone varied slightly between participants due to limited real-time position correction. Bretin et al. \cite{Bretin2023} investigated proximity, stress and discomfort in human-drone interaction in real and virtual environments. They conducted user studies with 42 participants comparing the same drone motion in real life and in virtual reality. They studied three movements - static far, approach, and static close, and asked participants to rate their discomfort levels for each. The results concluded that the approach phase was considered the most stressful while differing speeds did not affect comfort levels. Stress followed the same pattern through both testing methods, though each had different overall levels. They used a small DJI Tello drone (98 x 92.5 x 41 mm) and observed an increase in discomfort when entering the personal space (below 120 cm) thus suggesting a larger distance margin for comfort than the previous works. Yoon et al. \cite{Yoon2019} conducted a study where participants observed flying robots in virtual reality (VR) while their skin conductance was measured to understand their feelings of safety. Machine learning was used to estimate parameters for proposed safety measures during path planning. The flying robots were not allowed within the personal space region. They used HTC Vive VR headset in a custom Unity game engine-based VR test environment for human-aerial robot interactions and recruited 56 participants. They measured concurrent psychophysiological reactions of participants such as head motion kinematics and electrodermal activity (EDA) which were time-aligned with the velocity, altitude, and audio profile of the UAV's flight path. They suggested that the optimal trajectory is one generated using learned human models that can adapt to uncertain environments and human behaviours. Unlike in these works where the UAV approached in a single path, at a set altitude, with a fixed speed, and was autonomously controlled, in our research, we build on these existing work by taking a multipath approach and recruiting human pilots who added unpredictability to the flight making the participants more sensitive to risk. We also considered the case of the simultaneous multiple-robot approach.

Natural human interaction techniques of speech and visual gesture often require close proximity to succeed otherwise the originator may not be heard or seen by the receiver. Abioye et al. \cite{Abioye2022,abioye2019multimodal} proposed the multimodal speech and visual gesture (mSVG) human-aerial robot interaction technique to emulate human-to-human interaction for future aerial robot co-located with multiple humans where tangible control interfaces are not intuitive or practical. Cauchard et al. \cite{Cauchard2019} proposed the Drone.io gestural and visual interface for human-drone interaction to promote natural gesture interaction. Both of these techniques rely on the aerial robot maintaining proximity to the human to capture the human gesture or record speech. Natural human interaction techniques provide an alternative for human-swarm interaction when robots are co-located with humans in the same environment. However, there is a need to define safe operating zones for swarms. Therefore, in this research, we focus on determining safe operating zones for aerial swarms that minimise discomfort to the co-located human.

According to the literature, many factors could be responsible for discomfort. For example, Mesquita et al. \cite{Mesquita2022} investigated the impact of drones on wildlife by studying drone noise. The study revealed varied sensitivity amongst species, with Asian elephants showing heightened discomfort. While human responses may differ from those of animals, this natural response is critical to consider when discerning causes of discomfort. Ramos-Romero et al. \cite{Ramos-Romero2022} investigated requirements for drone operations to minimise the discomfort impacts of drone noise on people. They proposed a drone-fa\c cade distance, to meet WHO recommendations for sleep quality, from 110.6 m (slow flyover) to 179.1 m (fast flyover) for a large Gryphon Dynamics GD28X heavy lift Octocopter UAV with 2.1 m diameter capable of lifting about 32 kg, which was found to generate noise of up to 72.3 dB when hovering away at 50 m altitude. They suggested a link between drone noise, discomfort, and proximity. Raoult et al. \cite{Raoult2020} investigated operational protocols for the use of drones in marine animal research. They consider the effect of noise and wind generated by the drone propellers could cause physical disturbances to marine life. This discomfort could cause anxiety and trigger defensive behaviours. Therefore, in addition to noise, the wind generated by the UAV propeller is another component that can cause discomfort to co-located humans. MacArthur et al. \cite{MacArthur2017} investigated proximity and speed for social robots being deployed in healthcare, entertainment, and education. They found that in addition to proximity, the robot's speed of approach also caused discomfort. A slower-moving robot caused less discomfort than a fast-moving robot, especially when approaching a human. In Chang et al. \cite{chang_spiders_2017} user perception of drones, privacy, and security, the authors highlighted other factors that could contribute to discomforts such as the drone design, size, and colour, in addition to noise and wind, previously identified.

\section{Methodology}
\subsection{Experiment setup}
We developed a custom multirotor UAV using commercial-of-the-shelf components such as Hexsoon Edu 450 V2 drone frame kit, a Pixhawk-based flight controller (Cube Orange+), Here 3+ GNSS module, RadioMaster TX16S MKII transmitter, RadioMaster RP3 V2 ELRS receiver, and a Overlander 5000 mAh 14.8V 4S Lipo battery. The flight controller was flashed with the latest Ardupilot Copter firmware version 4.5.3. For the experiments, the UAVs were flown manually in stabilisation mode by two pilots with an average of 35 flight hours. The pilot had to pass Level 5 of the Real Flight Drone Simulator (RFDS) challenges. 

\begin{figure}[!htb]
    \centering
    \framebox{\includegraphics[width=0.99\columnwidth]{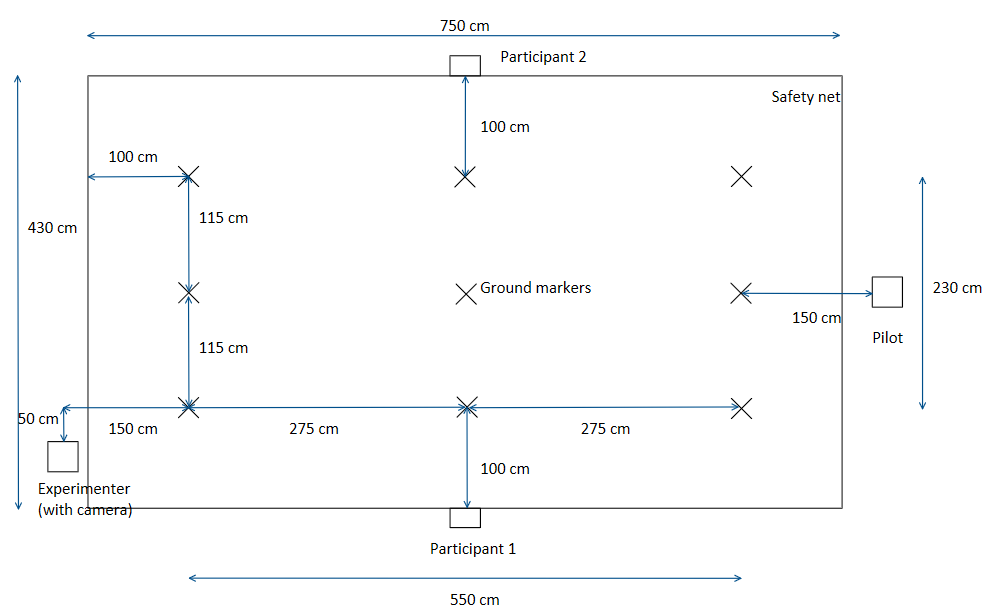}}
    \Description[Experiment setup layout]{A drawing showing the experiment setup layout with the position of participants, pilot, and the experimenter (observer)}
    \caption{Experiment setup layout}
    \label{fig:layout}
\end{figure}

The layout for the experiment is as shown in Figure \ref{fig:layout}. The flying arena was an area of dimension  750 x 430 x 600 cm, situated in the middle of the lab and bounded by safety nets on all sides and at the top. However, the flight area used for the experiment was 550 x 230 cm, allowing 100 cm of safety on all sides to prevent UAV collisions with the net. Markers were placed on the ground to guide UAV flight navigation so that UAV point-to-point trajectories were repeatable. The UAVs were flown at about 150 cm height level from the ground. The entire experiment consisted of six flight paths which add up to cover the flight area. A tally table was used to ensure each path was travelled at least three times overall and in both directions. An extra path was added for the multi-UAV test. The participants were outside the net, 100 cm from the closest marker, with both the pilot and experimenter (observer) positioned to the sides as shown in Figure \ref{fig:layout}. Participant 1 and Participant 2 labels in Figure \ref{fig:layout} mark the participant positions for experiment setup A and B respectively. The observer recorded a video of the drone movement which was post-processed to determine when each drone reached a marker and hence at what point during the path navigation the user pressed the discomfort button. To adhere to laboratory safety protocols, the UAV was flown within the safety net, with the participants, pilots, and observer (experimenter) outside the netted area. Participants were as close to a marker as safe flight allowed, 100 cm. Although the presence of the net was considered to potentially affect the study since the participants could have an enhanced feeling of safety outside the net, it was a safety requirement from our risk assessment and ethics approval. Also, all persons (participant, pilot, and observer) were required to wear safety goggles during flight experiments to comply with the laboratory safety protocol.

A button-based system recorded participant discomfort, ensuring a non-intrusive approach that collected objective responses. A timer code was developed in C programming language to record the time the participant pressed the spacebar on a laptop, indicating discomfort. Upon completing each flight, the code automatically saved these timestamps as a text file for easy data processing. These recorded timestamps were synchronised with the UAV's positional data to identify uncomfortable moments.

\subsection{Participants}
We recruited 18 participants for the experiment (11 males, 6 females, and 1 non-binary). The recruited participants were undergraduate students (with engineering and computer science backgrounds) in the 18-24 years age group. Table \ref{tab:participant_experiments} shows the distribution of the participants to each experiment condition. The last five participants (14 - 18) experienced both the single-drone and multiple-drone experiments. Due to inconsistent flight paths during participants 2, 3, 10, and 11 experiments, their data was excluded from the final result analysis.

\begin{table}[!htb]
\centering
\small
\renewcommand{\arraystretch}{1.3}
\caption{Participants experiment distribution.}
\begin{tabular}{|c|c|p{2.5cm}|c|}
\hline
Experiment & Setup & Participants  & No. of Participants \\
\hline
\multirow{2}{*}{Single Drone} & A & 1, 2, 3, 4, 6, 8, 10, 12, 14, 16, 18 & 11 \\
\cline{2-4}
 & B & 5, 7, 9, 11, 13, 15, 17 & 7 \\ 
\hline
\multirow{2}{*}{Multiple Drone} & A & 14, 16, 18 & 3 \\
\cline{2-4}
 & B &  15, 17 & 2 \\ 
\hline
\end{tabular}
\Description[Participants experiment distribution]{A table showing the participants' experiment distribution for the single and multiple drone experiments.}
\label{tab:participant_experiments}
\end{table}

\subsection{User study design} 
\begin{figure}[!htb]
    \centering
    \begin{subfigure}[c]{0.38\textwidth}
        \centering
        \framebox{\includegraphics[width=0.99\textwidth]{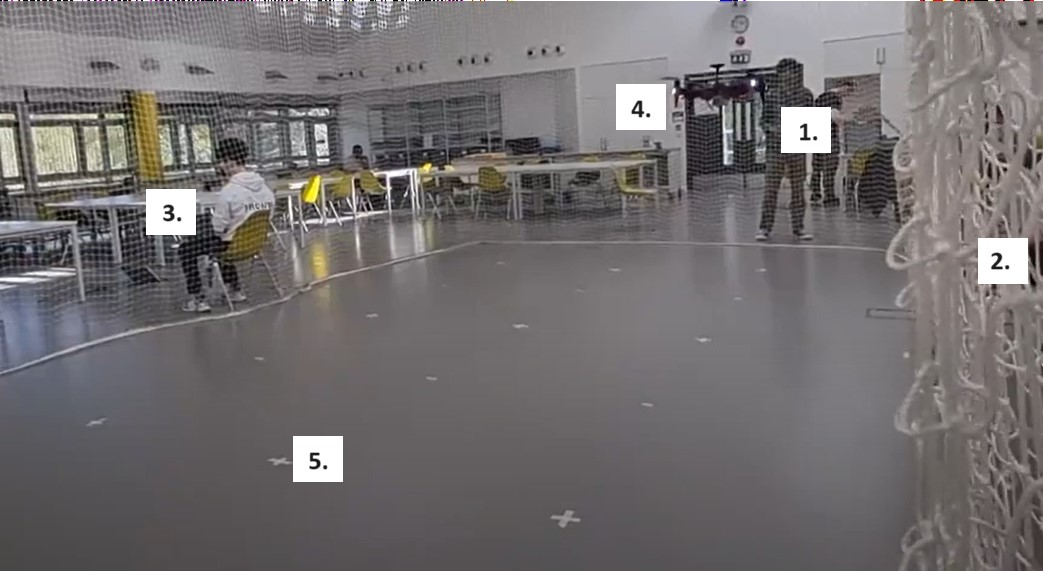}}
        \caption{Experiment 1 - single drone experiment \newline}
        \label{fig:experiment1}
    \end{subfigure}
    \begin{subfigure}[c]{0.45\textwidth}
        \centering
        \framebox{\includegraphics[width=0.99\textwidth]{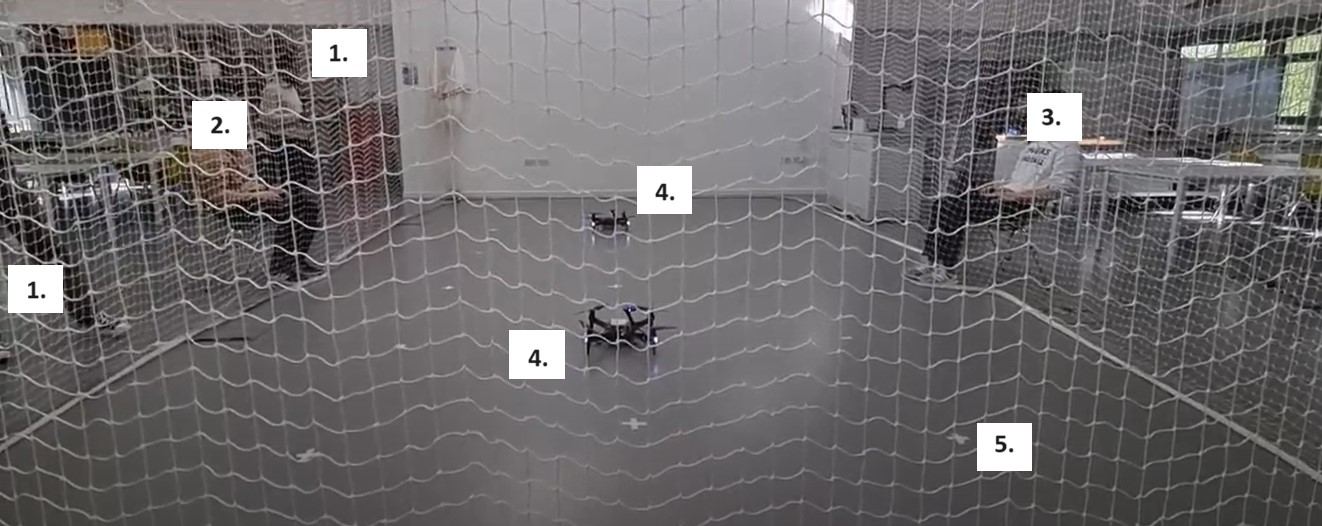}}
        \caption{Experiment 2 - multiple drone experiment}
        \label{fig:experiment2}
    \end{subfigure}
    \Description[Conducting the experiments]{A photograph showing the participant, drone pilots, and the experiment area during the experiment.}
    \caption{Conducting the experiments. Labels: (1) Pilots, (2) Observer, (3) Participants, and (4) Drones. }
    \label{fig:conducting_experiment}
\end{figure}

The user study consisted of two experiments, the single-drone and multiple-drone experiments as shown in Table \ref{tab:participant_experiments}. Each of the experiments consists of two setups based on the position of the participant during the experiment as shown in Figure \ref{fig:layout}. Participants were randomly assigned to each setup. The first 13 participants recruited conducted only the single-drone experiments. The last 5 participants recruited conducted both the single-drone and multiple-drone experiments. The first experiment was conducted to identify comfort zones between human and single UAV interactions. The second experiment was conducted to investigate the comfort zones for human-swarm interaction. In order to obtain practical measurements applicable to the real world, the user study was designed to be performed in the laboratory using real UAVs as shown in Figure \ref{fig:conducting_experiment}. The UAV flight trajectory consisted of multiple paths approaching and leaving the participant at different angles. The UAVs were flown manually due to the challenge of using GPS navigation in indoor environments and the technical limitations of setting up an alternative indoor positioning system. For the multiple drone experiments, 
the maximum number of drones we could fly manually within the flying arena was limited to two due to safety considerations and the size of the UAVs. Two pilots were required to operate each UAV individually.

The participants were seated 100 cm away from the closest UAV navigation point with a net barrier between them and the UAV and wearing safety goggles. Participants' discomfort was measured with the aid of a discomfort button pressed every time the participants felt uncomfortable. A 2D map of unsafe zones was generated based on the aggregate of discomfort button presses by all participants. In order to compare participants' comfort levels when facing the UAV and when facing away from the UAV, the first path of the experiment was repeated but with the participants facing away from the UAV and the data logged as in the initial test. Participants were asked to complete a brief post-study survey of their experience at the end of the experiment. We received ethics approval from the University of Southampton's ethics committee (confirmation number: ERGO/FEPS/91480).

\subsection{Procedure}
The experiment procedure is described below:
\begin{enumerate}
    \item[(a)] Participant briefing, consent, and setup (5 minutes)
    \item[(b)] Pre-experiment checks which include checking drone frame integrity, battery level, pilots are ready to fly, the observer is ready to record, and the button program is queued. (3 minutes)
    \item[(c)] Path 1 experiment: The pilot flies the UAV along the path 1 experiment trajectory, the observer records the flight trajectory, and participants press the discomfort button when they feel uncomfortable. The path 1 experiment finishes when the pilot completes the trajectory and lands the drone, the observer then stops the recording, and the button program is stopped and automatically saved.  (3 minutes) \label{procedure_path1_exp}
    \item[(d)] Path 2 experiment: The pilot flies the UAV along the path 2 experiment trajectory and everything else in Procedure \ref{procedure_path1_exp}c is repeated. (3 minutes)
    \item[(e)] Path 3 experiment: The pilot flies the UAV along the path 3 experiment trajectory and everything else in Procedure \ref{procedure_path1_exp}c is repeated. (3 minutes)
    \item[(f)] Path 4 experiment: The pilot flies the UAV along the path 4 experiment trajectory and everything else in Procedure \ref{procedure_path1_exp}c is repeated. (3 minutes)
    \item[(g)] Path 5 experiment: The pilot flies the UAV along the path 5 experiment trajectory and everything else in Procedure \ref{procedure_path1_exp}c is repeated. (3 minutes)
    \item[(h)] Path 6 experiment: The pilot flies the UAV along the path 6 experiment trajectory and everything else in Procedure \ref{procedure_path1_exp}c is repeated. (3 minutes)
    \item[(i)] Path 7 experiment: This is a repeat of the path 1 experiment but with the participants facing away from the UAV. The pilot flies the UAV along the path 1 experiment trajectory and everything else in Procedure \ref{procedure_path1_exp}c is repeated. (3 minutes)
    \item[(j)] Path 8 experiment (for Participants 14 - 18 only): This is the multiple drone experiment. The two pilots fly their drones along their specific UAV trajectory, the observer records the flight trajectory, and participants press the discomfort button when they feel uncomfortable. The path 8 experiment finishes when the two pilots complete the trajectory and land the drones, the observer then stops the recording, and the button program is stopped and automatically saved. (3 minutes)
    \item[(k)] Post-study survey and study conclusion: the participants are then asked to complete the post-study survey which asked the participants (i) if being unable to see the UAV made them feel uncomfortable, (ii) what factor contributed to their discomfort the most, and (iii) which of the experiments (single drone or multiple drone) caused the most discomfort? After this, the participants may leave. (5 minutes)
\end{enumerate}

\begin{figure*}[t]
    \centering
    \begin{subfigure}[b]{0.33\textwidth}
        \centering
        \includegraphics[width=0.99\textwidth]{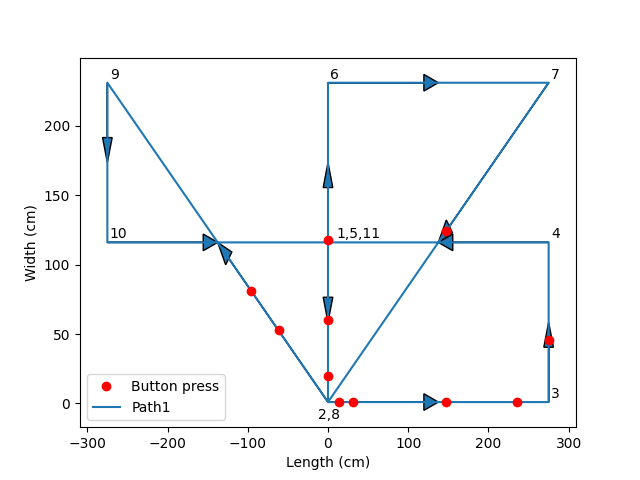}
        \caption{Path1 Position}
        \label{fig:forward_v4_path1_position}
    \end{subfigure}    
    \hfill
    \begin{subfigure}[b]{0.33\textwidth}
        \centering
        \includegraphics[width=0.99\textwidth]{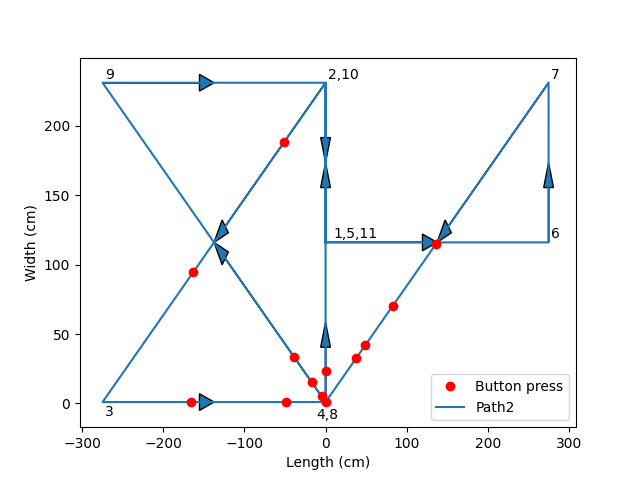}
        \caption{Path2 Position}
        \label{fig:forward_v4_path2_position}
    \end{subfigure}    
    \hfill
    \begin{subfigure}[b]{0.33\textwidth}
        \centering
        \includegraphics[width=0.99\textwidth]{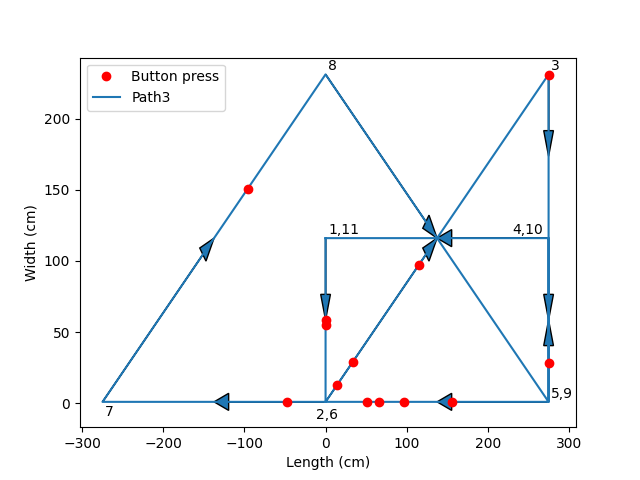}
        \caption{Path3 Position}
        \label{fig:forward_v4_path3_position}
    \end{subfigure}    
    \hfill
    \begin{subfigure}[b]{0.33\textwidth}
        \centering
        \includegraphics[width=0.99\textwidth]{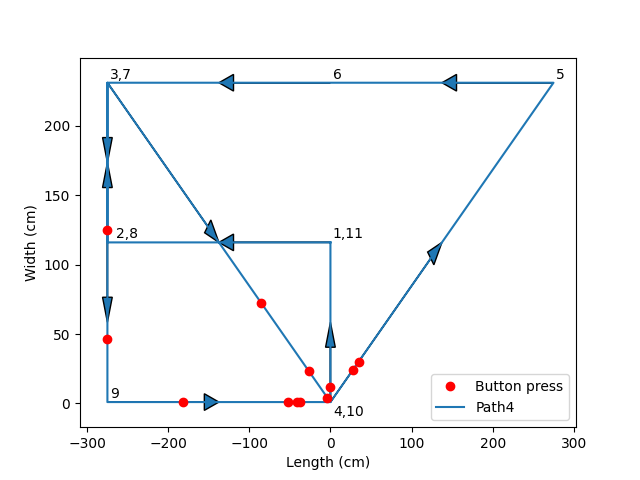}
        \caption{Path4 Position}
        \label{fig:forward_v4_path4_position}
    \end{subfigure}
    \hfill
    \begin{subfigure}[b]{0.33\textwidth}
        \centering
        \includegraphics[width=0.99\textwidth]{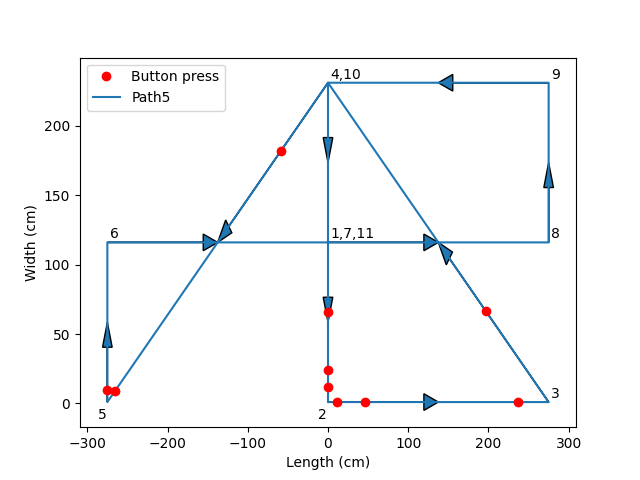}
        \caption{Path5 Position}
        \label{fig:forward_v4_path5_position}
    \end{subfigure}
    \hfill
    \begin{subfigure}[b]{0.33\textwidth}
        \centering
        \includegraphics[width=0.99\textwidth]{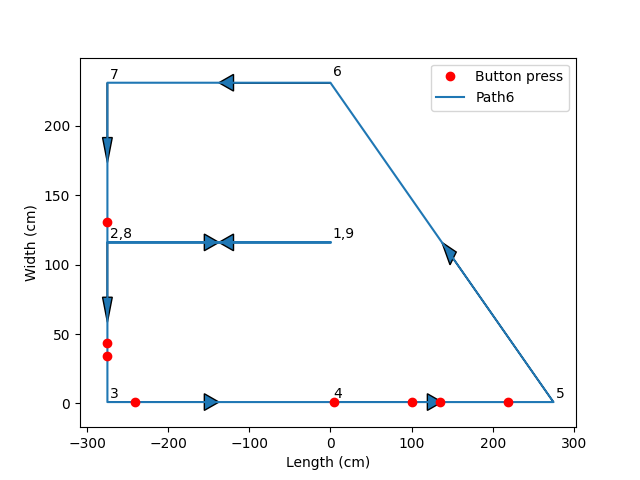}
        \caption{Path6 Position}
        \label{fig:forward_v4_path6_position}
    \end{subfigure}
    \Description[Experiment 1 Setup A path 1 -- 6 results]{Experiment 1 Setup A path 1 -- 6 results.}
    \caption{Experiment 1 Setup A path 1 -- 6 results. UAV navigation goes from 1 to 2, to 3, to 4 etc. as indicated by the numbers at the ends of each line segment, finishing at 11 (or 9 for Path6). Blue arrows show the direction of UAV travel. Red dots indicate the position of the UAV when the discomfort button is pressed.}
    \label{fig:forward_path_result}
\end{figure*}

Each participant's experiment lasted about 40 minutes. At the end of the experiments, the button data file is checked to ensure that the participant's button press data were logged correctly, the video recorded by the observer is checked to ensure that the flight paths are properly captured for post-processing, the UAV frames and electronics are checked for tight couplings and connections, and the LiPo battery removed for recharge in preparation for the next participant experiments.

\section{Results}
The results are presented in three parts. The first part discusses the result of Experiment 1 Setup A, the second part discusses Experiment 1 Setup B, and the last part discusses Experiment 2 Setup A and B. In performing the analysis of the button press data, the button press per participant was limited to only one per flight path segment. In cases where the participant pressed the button multiple times within a flight path segment, the average was computed and used. This was to ensure multiple button presses per segment were due to multiple unique participants indicating discomfort.

\subsection{Experiment 1 Setup A (Single Drone)}

Figure \ref{fig:forward_path_result} shows the Experiment 1A 2D navigation plot of the UAV flight trajectory for each of the 6 flight paths in Blue. The Blue arrows in the middle of each line segment indicate the UAV's direction of travel. The Red dots indicate the point during the flight when the participants pressed the discomfort button. As can be seen from the plots, there is a concentration of Red dots closer to the participant, indicating increased discomfort when the UAV is close to the participants.

\begin{figure*}[t]
    \centering
    \begin{subfigure}[b]{0.48\textwidth}
        \centering
        \includegraphics[width=0.99\textwidth]{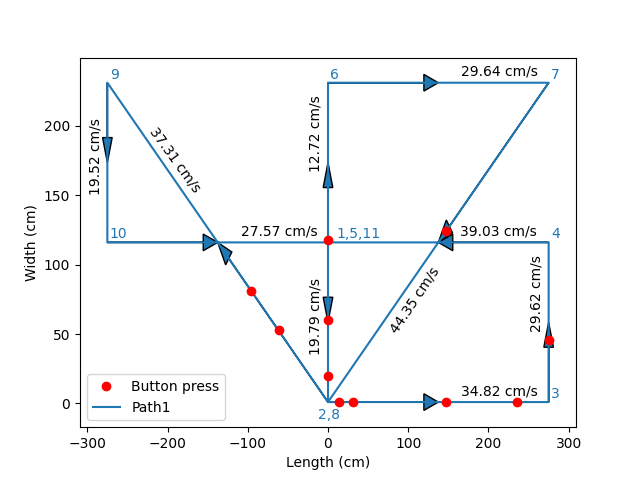}
        \caption{Path1 facing drone test}
        \label{fig:forward_v4_path1_velocity}
    \end{subfigure}
    \hfill
    \begin{subfigure}[b]{0.48\textwidth}
        \centering
        \includegraphics[width=0.99\textwidth]{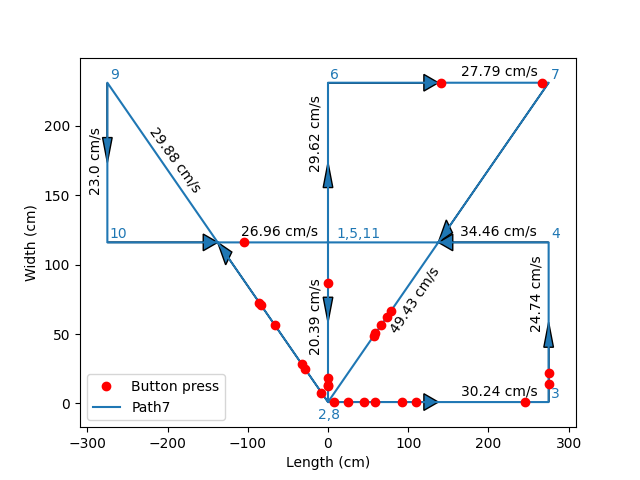}
        \caption{Path7 (Path1 backing drone) blindfold test}
        \label{fig:forward_v4_path7_velocity}
    \end{subfigure}
    \Description[Experiment 1A path blindfold test results]{Experiment 1A path blindfold test results.}
    \caption{Experiment 1A path blindfold test results. Blue arrows show the direction of UAV travel. Red dots indicate the position of the UAV when the discomfort button is pressed. UAV navigation goes from 1 to 2, to 3, to 4 etc. as indicated by the numbers at the ends of each line segment, finishing at 11. Text on the line segments shows flight segment velocities in cm/s.}
    \label{fig:forward_path_result_blindtest}
\end{figure*}

Figure \ref{fig:forward_path_result_blindtest} shows the plot of the path 1 experiment when the participant is facing forward (the drone) and path 7 which is a repeat of the path 1 experiment but with the participant facing backwards (away from the drone). It can be observed that for the same path experienced by the same participants, there is an increase in the number of times the discomfort button was pressed when the participant was not seeing the drone. The discomfort button was pressed 11 times in the path 1 experiment and 27 times in the path 7 experiment. That is about 2.5 times more. This shows that the participants were uncomfortable with a drone approach from behind and were 2.5 times more likely to be stressed. Figure \ref{fig:forward_path_result_blindtest} also showed the average segment flight velocity in cm/s. The flight velocity for segments 1-2 was 19.79 cm/s in the path 1 facing experiment and 20.39 cm/s in the path 1 backing experiment (path 7). The difference in segment velocities was due to the UAV being flown manually by a human pilot. The most same-segment difference was observed in segment 5-6 where the path 1 segment velocity was 12.72 cm/s and the path 7 (path 1 backing) segment velocity was 29.62 cm/s.

\begin{figure*}[t]
    \centering
    \begin{subfigure}[b]{0.33\textwidth}
        \centering
        \includegraphics[width=0.99\textwidth]{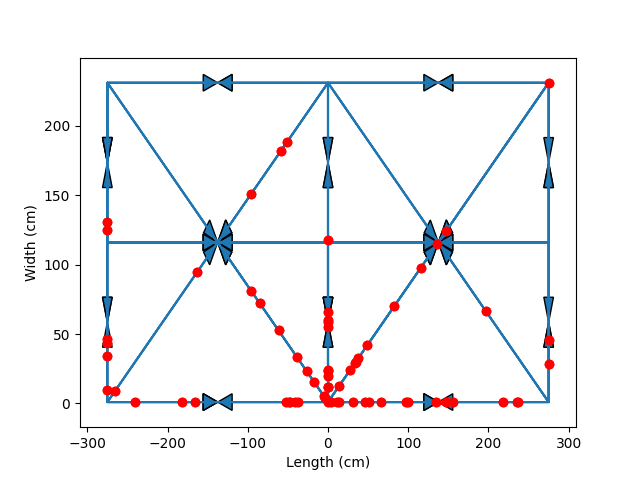}
        \caption{Experiment 1A combined paths}
        \label{fig:forward_v4_combined_path_position}
    \end{subfigure}
    \hfill
    \begin{subfigure}[b]{0.33\textwidth}
        \centering
        \includegraphics[width=0.99\textwidth]{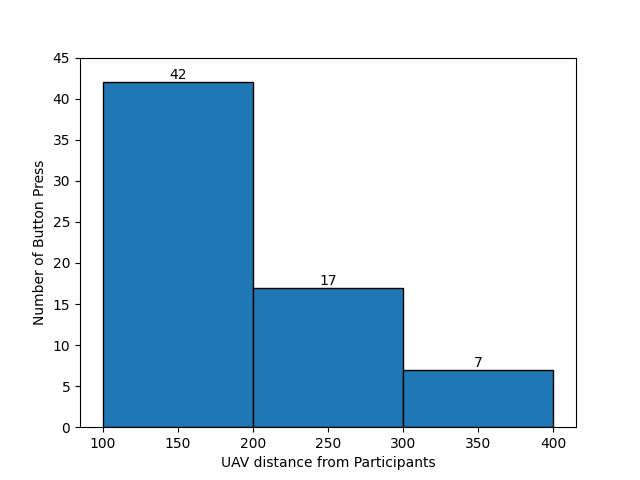}
        \caption{Histogram distribution of distance}
        \label{fig:forward_v4_combined_path_position_histogram}
    \end{subfigure}
    \hfill
    \begin{subfigure}[b]{0.33\textwidth}
        \centering
        \includegraphics[width=0.99\textwidth]{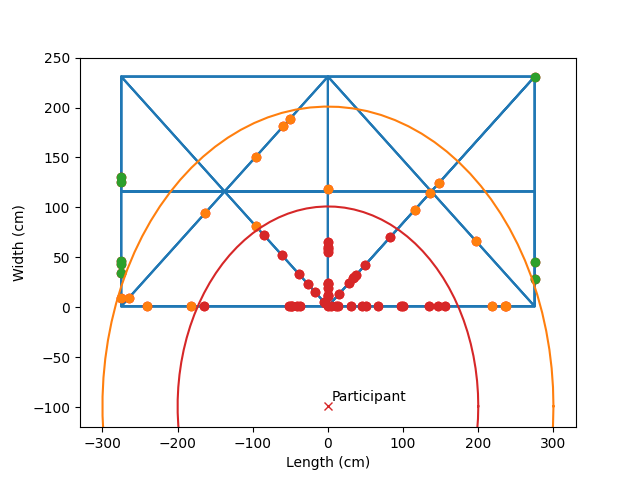}
        \caption{Segmenting histogram range}
        \label{fig:forward_v4_combined_path_position_ranged}
    \end{subfigure}
    \Description[Combined flight paths plots, histograms and radial segments plots]{Combined flight paths plots, histograms of discomfort button press distances, and radial segments highlight safe flight zones plots.}
    \caption{UAV comfortable distance zoning using histogram and segment plots}
    \label{fig:forward_path_combined_result}
\end{figure*}

Figure \ref{fig:forward_v4_combined_path_position} shows the combined path 1 to 6 plots overlaid on each other. The button presses of each path were also overlaid so the distribution of where the participants were uncomfortable with the drone presence could be mapped. From this Figure, it can be observed that the participants were generally uncomfortable when the drone was nearest to them. Figure \ref{fig:forward_v4_combined_path_position_histogram} shows the histogram distribution of the participants' button presses. The histogram plot started at 100 cm because there was a 100 cm safety buffer between the participants and the drone flight path. This means there were no points below 100 cm. Also, for easier quantisation, the histogram bins are at an interval of 100 cm. From the histogram plots, three zones emerged, the Red zone marking the point of most discomfort (100 - 200 cm), the Yellow zone marking points of moderately acceptable discomfort (200 - 300 cm),  and the Green zone marking points of little or no discomfort (> 300 cm). About 64 \% of the total button presses (66) occurred in the 100 - 200 cm zone, about 26 \% in the 200 - 300 cm zone, and the remaining 10 \% in the 300 - 400 cm zone. Figure \ref{fig:forward_v4_combined_path_position_ranged} shows the combined Experiment 1A path plots segmented by radial lines at distances of 100 cm in Red and 200 cm in Yellow colours with the button press points that fall within this being colour-coded to indicate which region they fall into.

\subsection{Experiment 1 Setup B (Single Drone)}

\begin{figure*}[t]
    \centering
    \begin{subfigure}[b]{0.33\textwidth}
        \centering
        \includegraphics[width=0.99\textwidth]{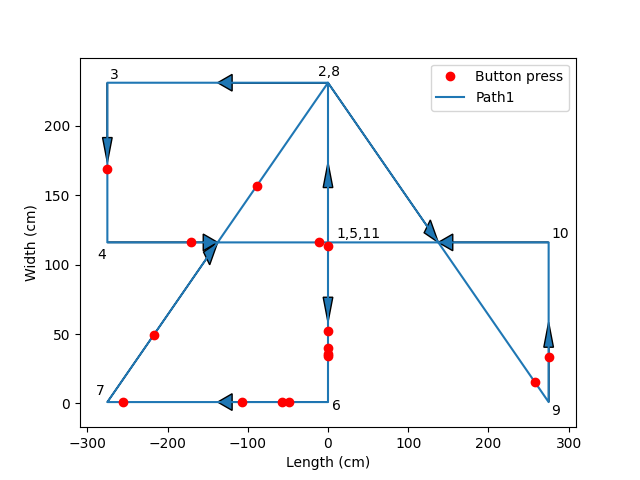}
        \caption{Path1 Position}
        \label{fig:reverse_v1_path1_position}
    \end{subfigure}
    \hfill
    \begin{subfigure}[b]{0.33\textwidth}
        \centering
        \includegraphics[width=0.99\textwidth]{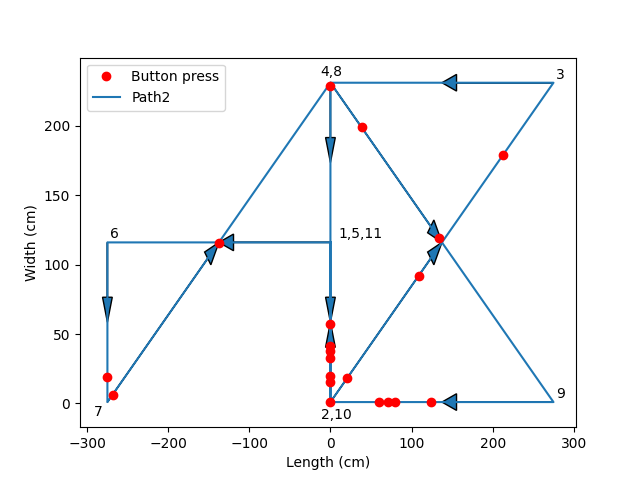}
        \caption{Path2 Position}
        \label{fig:reverse_v1_path2_position}
    \end{subfigure}
    \hfill
    \begin{subfigure}[b]{0.33\textwidth}
        \centering
        \includegraphics[width=0.99\textwidth]{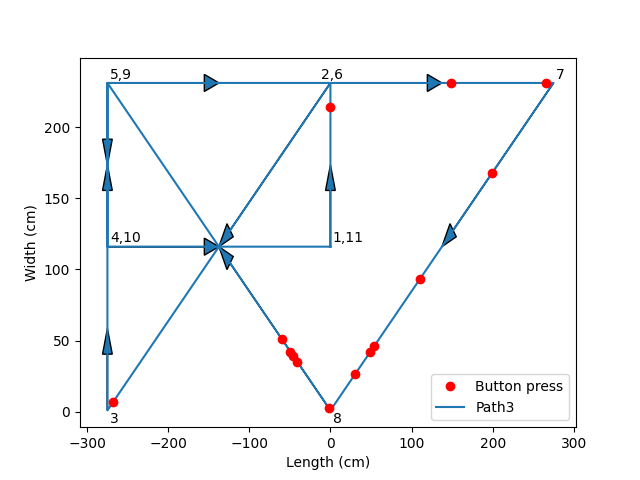}
        \caption{Path3 Position}
        \label{fig:reverse_v1_path3_position}
    \end{subfigure}
    \hfill
    \begin{subfigure}[b]{0.33\textwidth}
        \centering
        \includegraphics[width=0.99\textwidth]{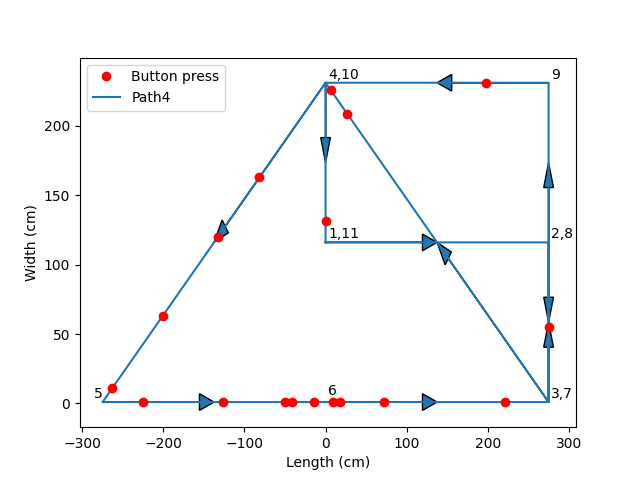}
        \caption{Path4 Position}
        \label{fig:reverse_v1_path4_position}
    \end{subfigure}
    \hfill
    \begin{subfigure}[b]{0.33\textwidth}
        \centering
        \includegraphics[width=0.99\textwidth]{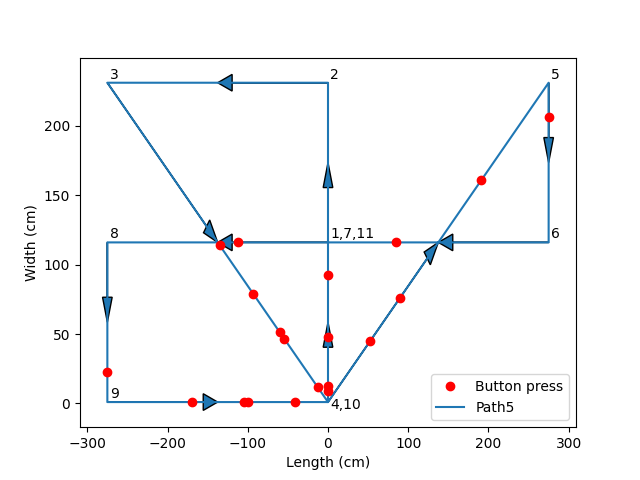}
        \caption{Path5 Position}
        \label{fig:reverse_v1_path5_position}
    \end{subfigure}
    \hfill
    \begin{subfigure}[b]{0.33\textwidth}
        \centering
        \includegraphics[width=0.99\textwidth]{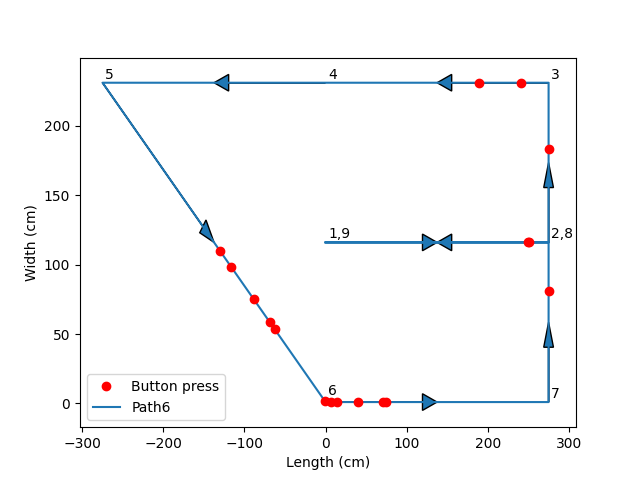}
        \caption{Path6 Position}
        \label{fig:reverse_v1_path6_position}
    \end{subfigure}
    \Description[Experiment 1 Setup B path 1 -- 6 results]{Experiment 1 Setup B path 1 -- 6 results.}
    \caption{Experiment 1 Setup B path 1 -- 6 results. UAV navigation goes from 1 to 2, to 3, to 4 etc. as indicated by the numbers at the ends of each line segment, finishing at 11 (or 9 for Path6). Blue arrows show the direction of UAV travel. Red dots indicate the position of the UAV when the discomfort button is pressed.}
    \label{fig:reverse_path_result}
\end{figure*}

Figure \ref{fig:reverse_path_result} shows the Experiment 1B 2D navigation plot of the UAV flight trajectory for each of the 6 flight paths in Blue. The Blue arrows in the middle of each line segment indicated the UAV's direction of travel. The Red dots indicated the point during the flight when the participants pressed the discomfort button. As can be seen from Figure \ref{fig:reverse_path_result}, there is a concentration of Red dots closer to the participant and towards the top right area, indicating multiple areas of increased discomfort, one of which was proximity to the participants.

\begin{figure*}[!htb]
    \centering
    \begin{subfigure}[b]{0.48\textwidth}
        \centering
        \includegraphics[width=0.99\textwidth]{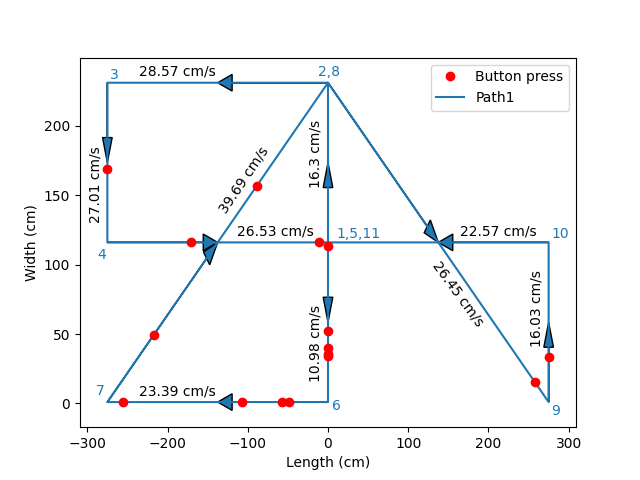}
        \caption{Path1 facing drone test}
        \label{fig:reverse_v1_path1_velocity}
    \end{subfigure}
    \hfill
    \begin{subfigure}[b]{0.48\textwidth}
        \centering
        \includegraphics[width=0.99\textwidth]{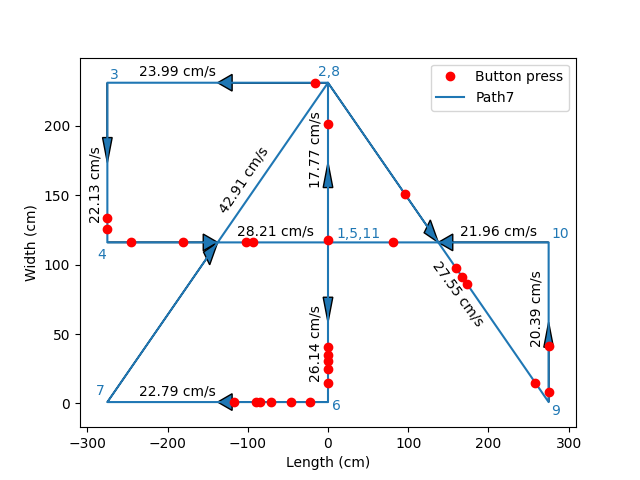}
        \caption{Path7 (Path1 backing drone) blindfold test}
        \label{fig:reverse_v1_path7_velocity}
    \end{subfigure}
    \Description[Experiment 1B path blindfold test results]{Experiment 1B path blindfold test results.}
    \caption{Experiment 1B path blindfold test results. Blue arrows show the direction of UAV travel. Red dots indicate the position of the UAV when the discomfort button is pressed. UAV navigation goes from 1 to 2, to 3, to 4 etc. as indicated by the numbers at the ends of each line segment, finishing at 11. Text on the line segments shows flight segment velocities in cm/s.}
    \label{fig:reverse_path_result_blindtest}
\end{figure*}

\begin{figure*}[t]    
    \centering
    \begin{subfigure}[b]{0.33\textwidth}
        \centering
        \includegraphics[width=0.99\textwidth]{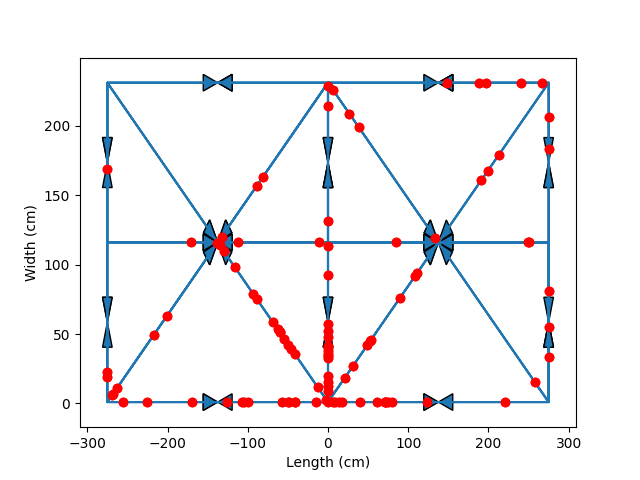}
        \caption{Experiment 1B combined paths}
        \label{fig:reverse_v1_combined_path_position}
    \end{subfigure}
    \hfill
    \begin{subfigure}[b]{0.33\textwidth}
        \centering
        \includegraphics[width=0.99\textwidth]{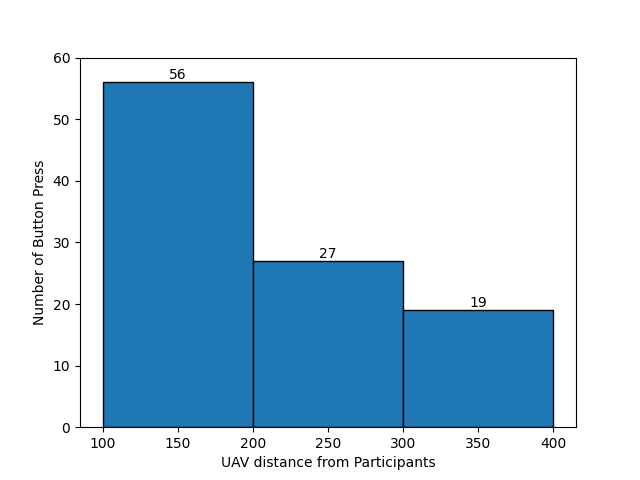}
        \caption{Histogram distribution of distances}
        \label{fig:reverse_v1_combined_path_position_histogram}
    \end{subfigure}
    \hfill
    \begin{subfigure}[b]{0.33\textwidth}
        \centering
        \includegraphics[width=0.99\textwidth]{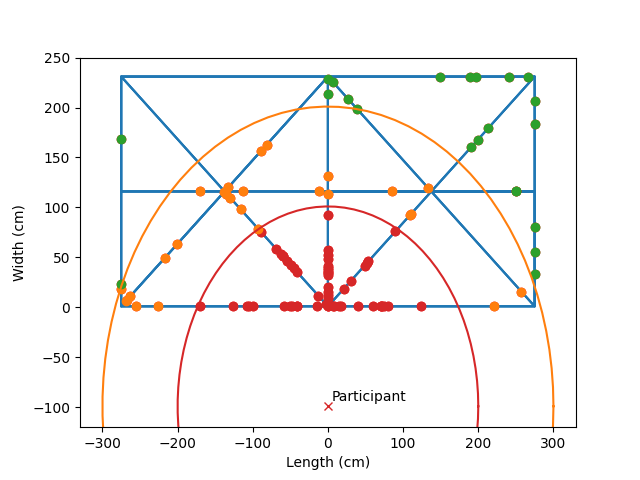}
        \caption{Segmenting histogram range}
        \label{fig:reverse_v1_combined_path_position_ranged}
    \end{subfigure}
    \Description[Experiment 1B combined flight paths plots, histograms and radial segments plots]{Combined flight paths plots, histograms of discomfort button press distances, and radial segments highlight safe flight zones plots for Experiments 1B.}
    \caption{UAV comfortable distance zoning using histogram and segment plots}
    \label{fig:reverse_path_combined_result}
\end{figure*}

Figure \ref{fig:reverse_path_result_blindtest} shows the plot of the path 1 experiment when the participant is facing forward (the drone) and path 7 which is a repeat of the path 1 experiment but with the participant facing backwards (away from the drone) for Experiment 1 Setup B. It can be observed that for the same path experienced by the same participants, there is an increase in the number of times the discomfort button was pressed when the participant was not seeing the drone. The discomfort button was pressed 16 times in the path 1 experiment and 28 times in the path 7 experiment. That is about 1.75 times more. This shows that the participants were uncomfortable with a drone approach from behind and were 1.75 times more likely to be stressed. Figure \ref{fig:reverse_path_result_blindtest} also showed the average segment flight velocity in cm/s. The most same-segment difference was observed in segment 5-6 where the path 1 segment velocity was 10.98 cm/s and the path 7 (path 1 backing) segment velocity was 26.14 cm/s.

Figure \ref{fig:reverse_v1_combined_path_position} shows the combined path 1 to 6 plots overlaid on each other for the Experiment 1B condition. The button presses of each path were also overlaid so the distribution of where the participants were uncomfortable with the drone presence could be mapped. From this Figure, it can be observed that the participants were generally uncomfortable when the drone was nearest to them. Also, there is a clear concentration of button press points at the top right-hand corner. This position coincided with the position of the observer and it seems that some of the participants were triggering the distress button out of concern or empathy for the observer (experimenter in Figure \ref{fig:layout}). Figure \ref{fig:reverse_v1_combined_path_position_histogram} shows the histogram distribution of the participants' button presses. The histogram plot started at 100 cm because there was a 100 cm safety buffer between the participants and the drone flight path. This means there were no points below 100 cm. Also, for easier quantisation, the histogram bins are at an interval of 100 cm. From the histogram plots, three zones emerged, the Red zone marking the point of most discomfort (100 - 200 cm), the Yellow zone marking points of moderately acceptable discomfort (200 - 300 cm),  and the Green zone marking points of little or no discomfort (> 300 cm). About 55 \% of the total button presses (102) occurred in the 100 - 200 cm zone, about 26 \% in the 200 - 300 cm zone, and the remaining 19 \% in the 300 - 400 cm zone. Figure \ref{fig:reverse_v1_combined_path_position_ranged} shows the combined Experiment 1A path plots segmented by radial lines at distances of 100 cm in Red and 200 cm in Yellow colours with the button press points that fall within this being colour-coded to indicate which region they fall into.

\subsection{Experiment 2 Setup A and B (Multiple Drones)}

\begin{figure*}[t]
    \centering
    \begin{subfigure}[b]{0.33\textwidth}
        \centering
        \includegraphics[width=0.99\textwidth]{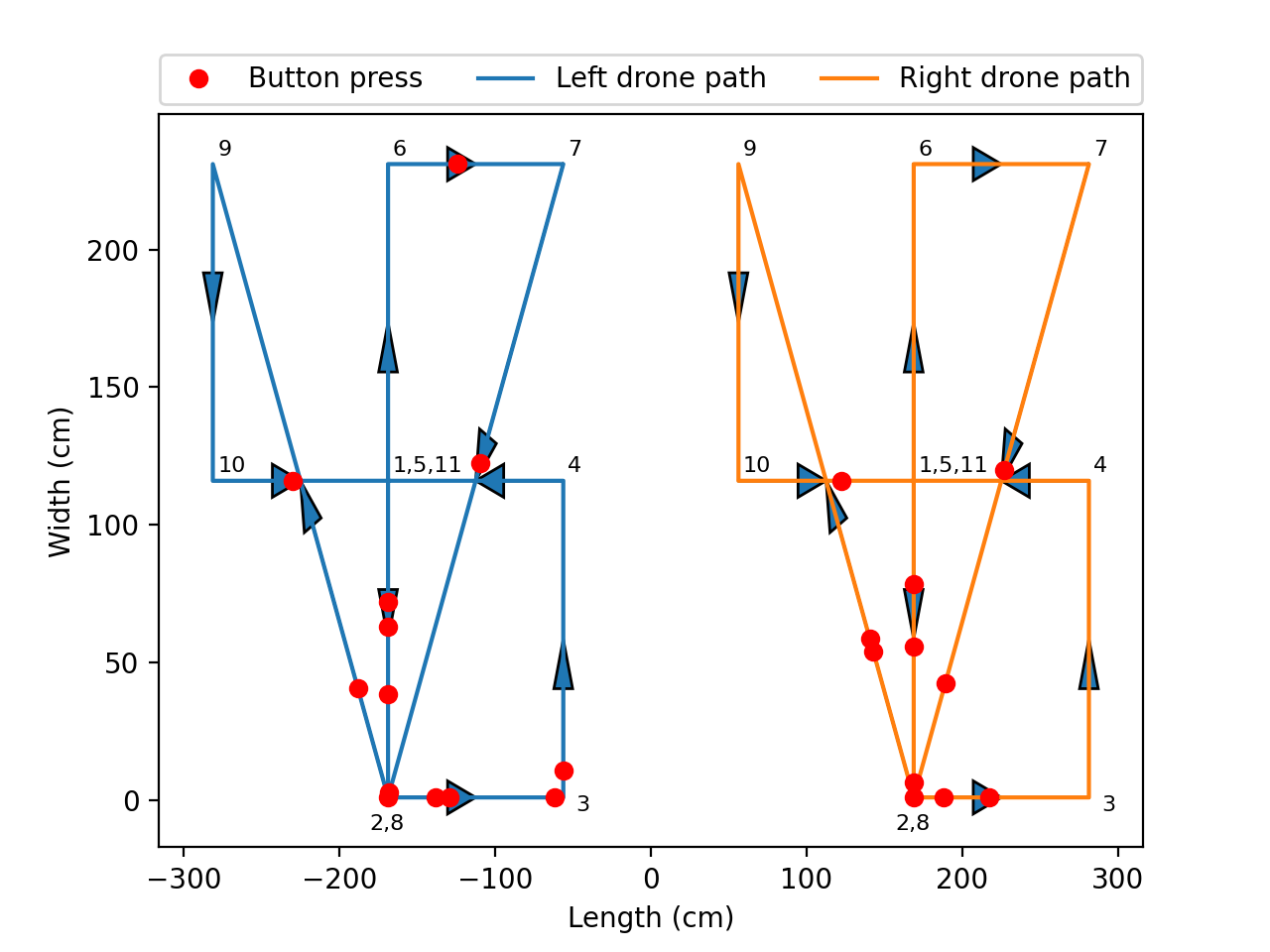}
        \caption{Multiple Drones Set A Position}
        \label{fig:swarm_fwd_path_position_v1}
    \end{subfigure}
    \hfill
    \begin{subfigure}[b]{0.33\textwidth}
        \centering
        \includegraphics[width=0.99\textwidth]{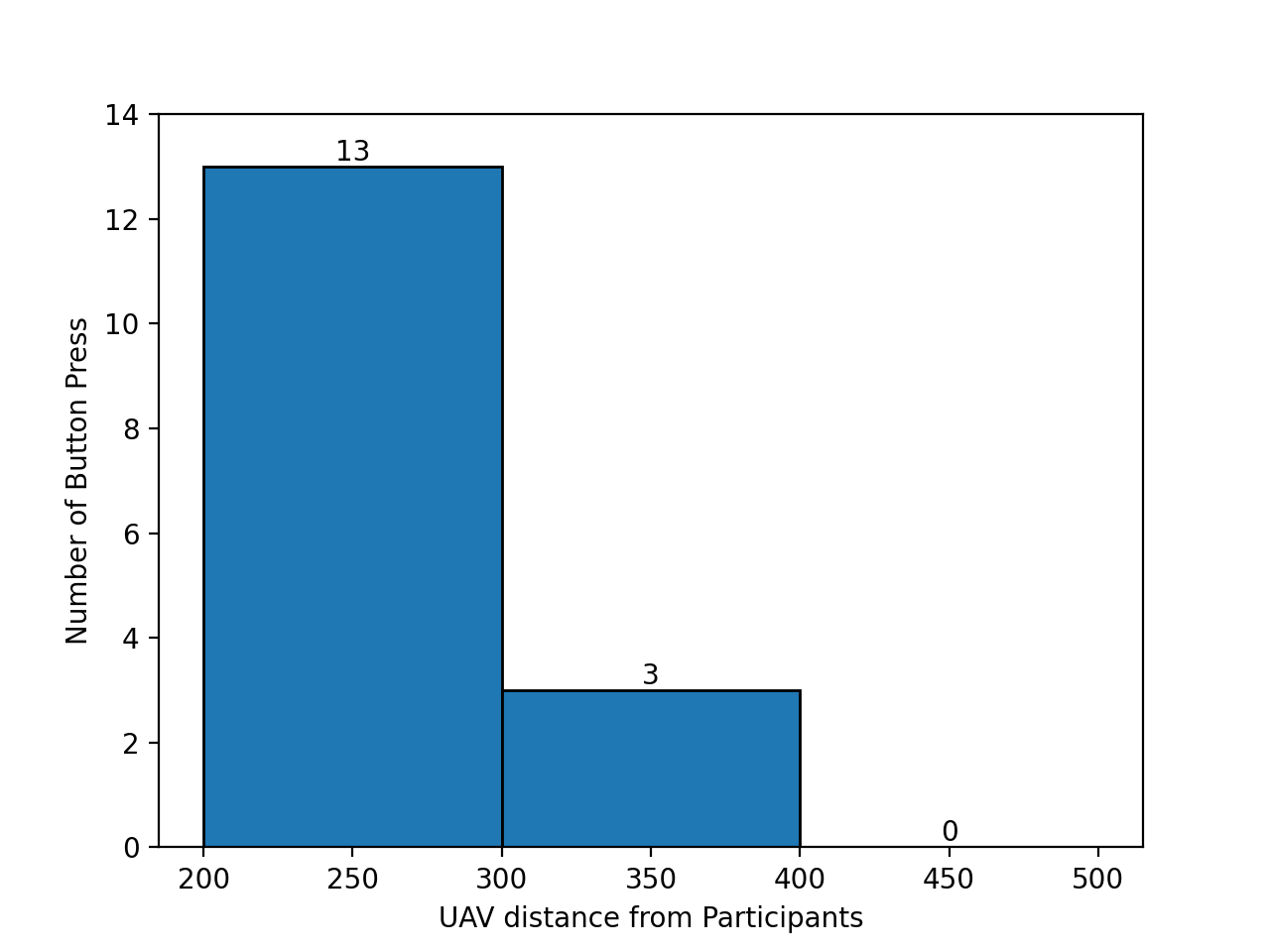}
        \caption{Setup A distance histogram}
        \label{fig:swarm_fwd_button_press_histogram_v1}
    \end{subfigure}
    \hfill
    \begin{subfigure}[b]{0.33\textwidth}
        \centering
        \includegraphics[width=0.99\textwidth]{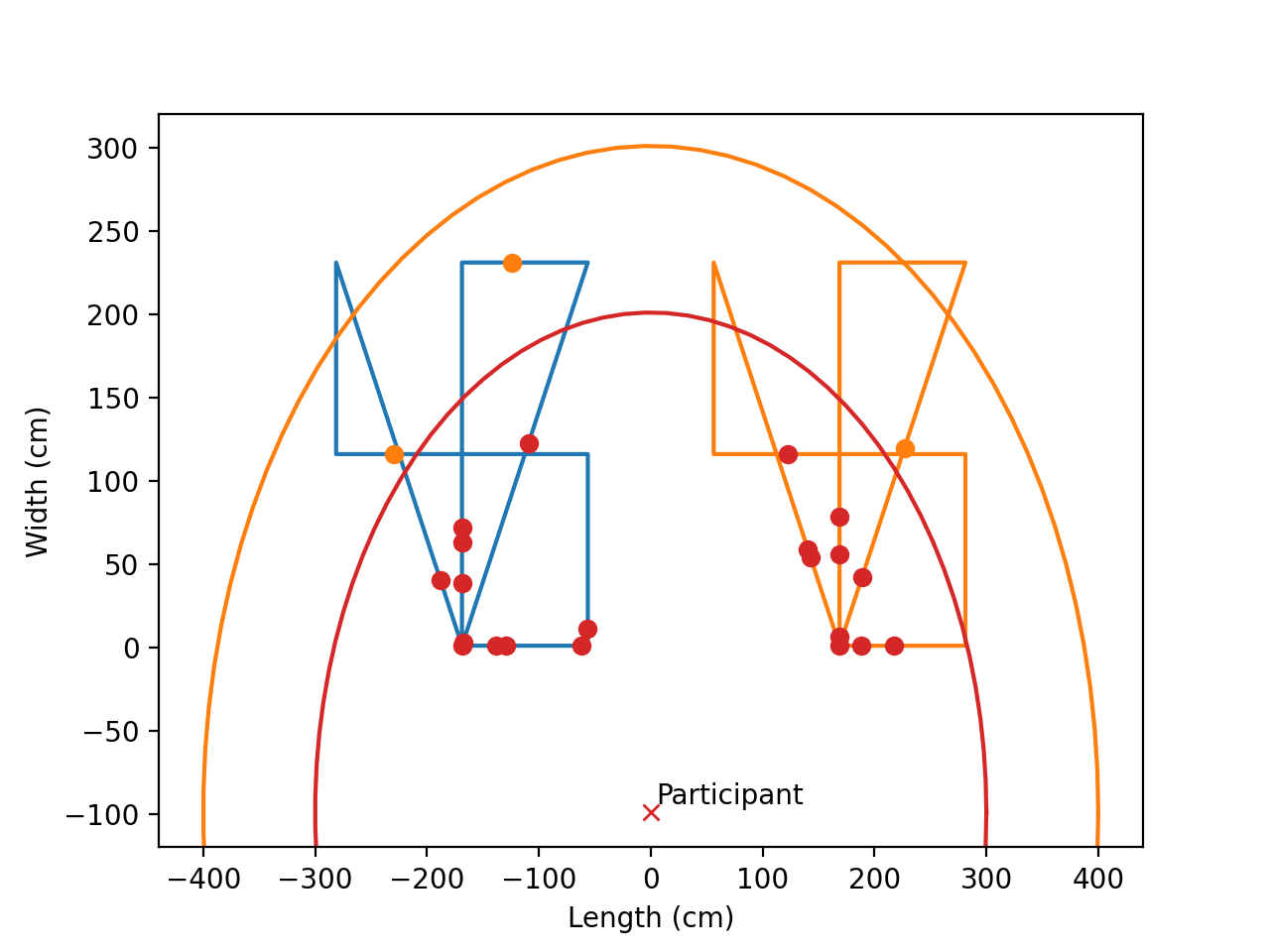}
        \caption{Setup A distance band region}
        \label{fig:swarm_fwd_path_ranged_v1}
    \end{subfigure}
    \hfill
    \begin{subfigure}[b]{0.33\textwidth}
        \centering
        \includegraphics[width=0.99\textwidth]{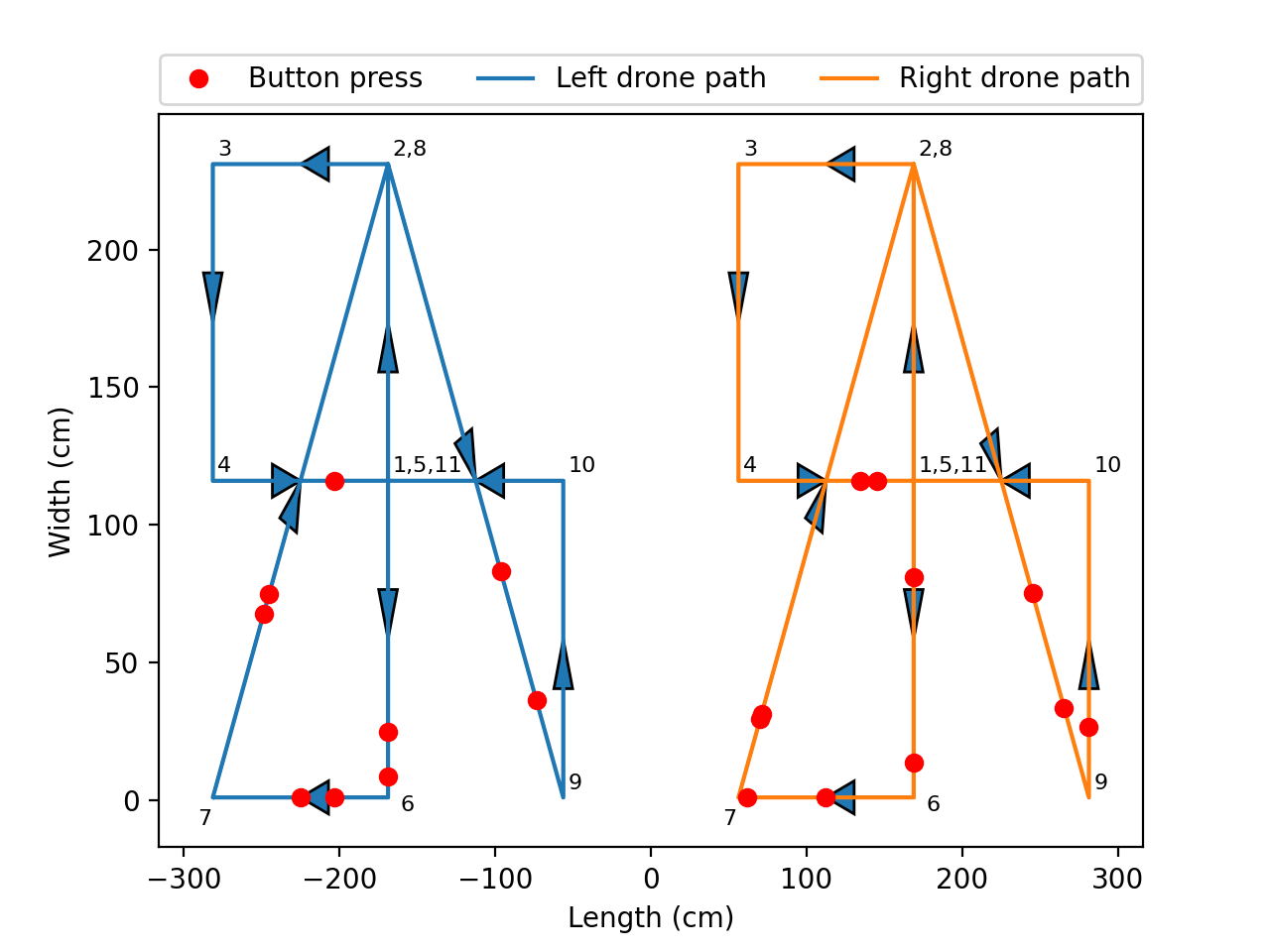}
        \caption{Multiple Drones Set B Position}
        \label{fig:swarm_rev_path_position_v1}
    \end{subfigure}
    \hfill
    \begin{subfigure}[b]{0.33\textwidth}
        \centering
        \includegraphics[width=0.99\textwidth]{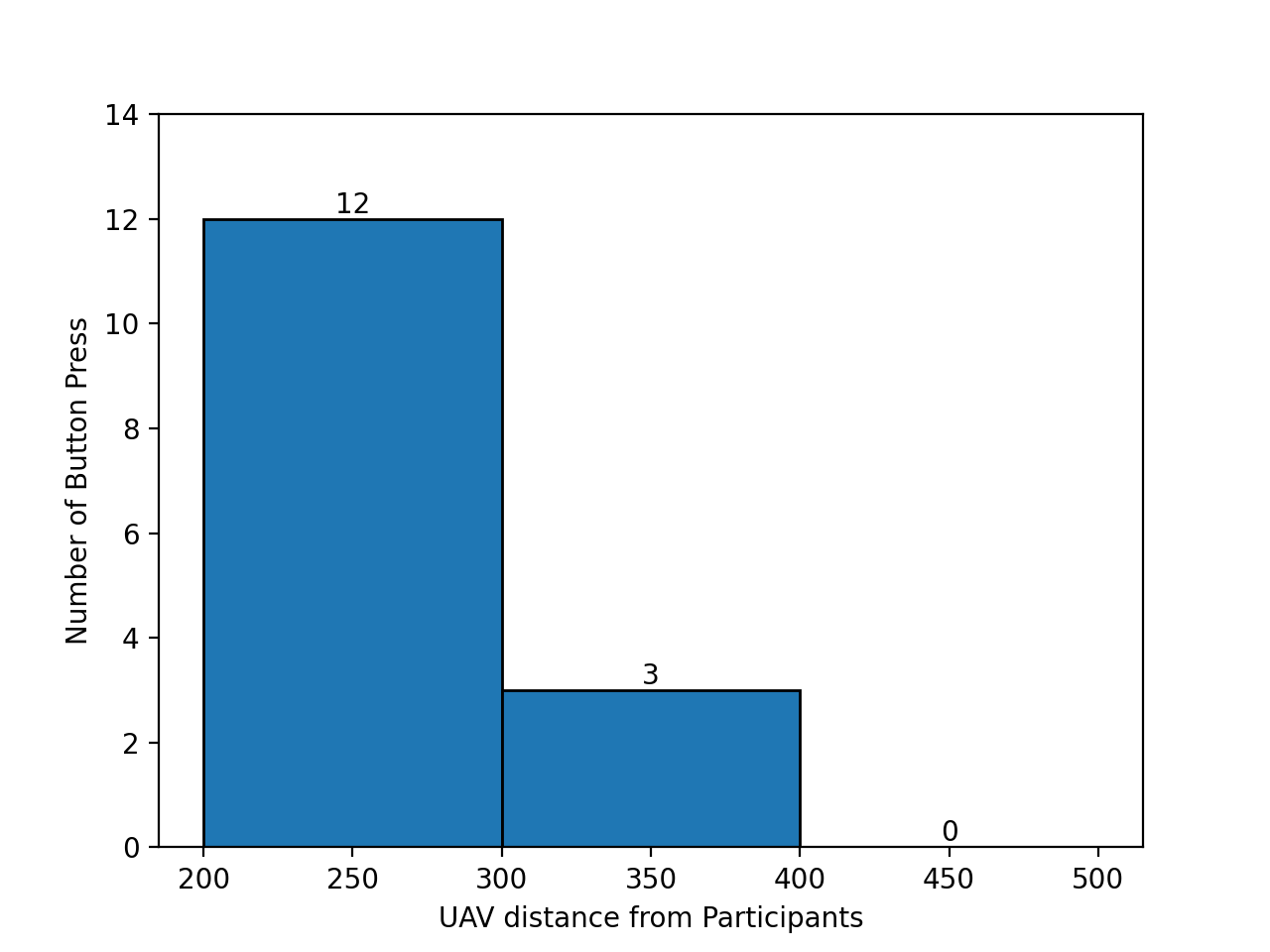}
        \caption{Setup B distance histogram}
        \label{fig:swarm_rev_button_press_histogram_v1}
    \end{subfigure}
    \hfill
    \begin{subfigure}[b]{0.33\textwidth}
        \centering
        \includegraphics[width=0.99\textwidth]{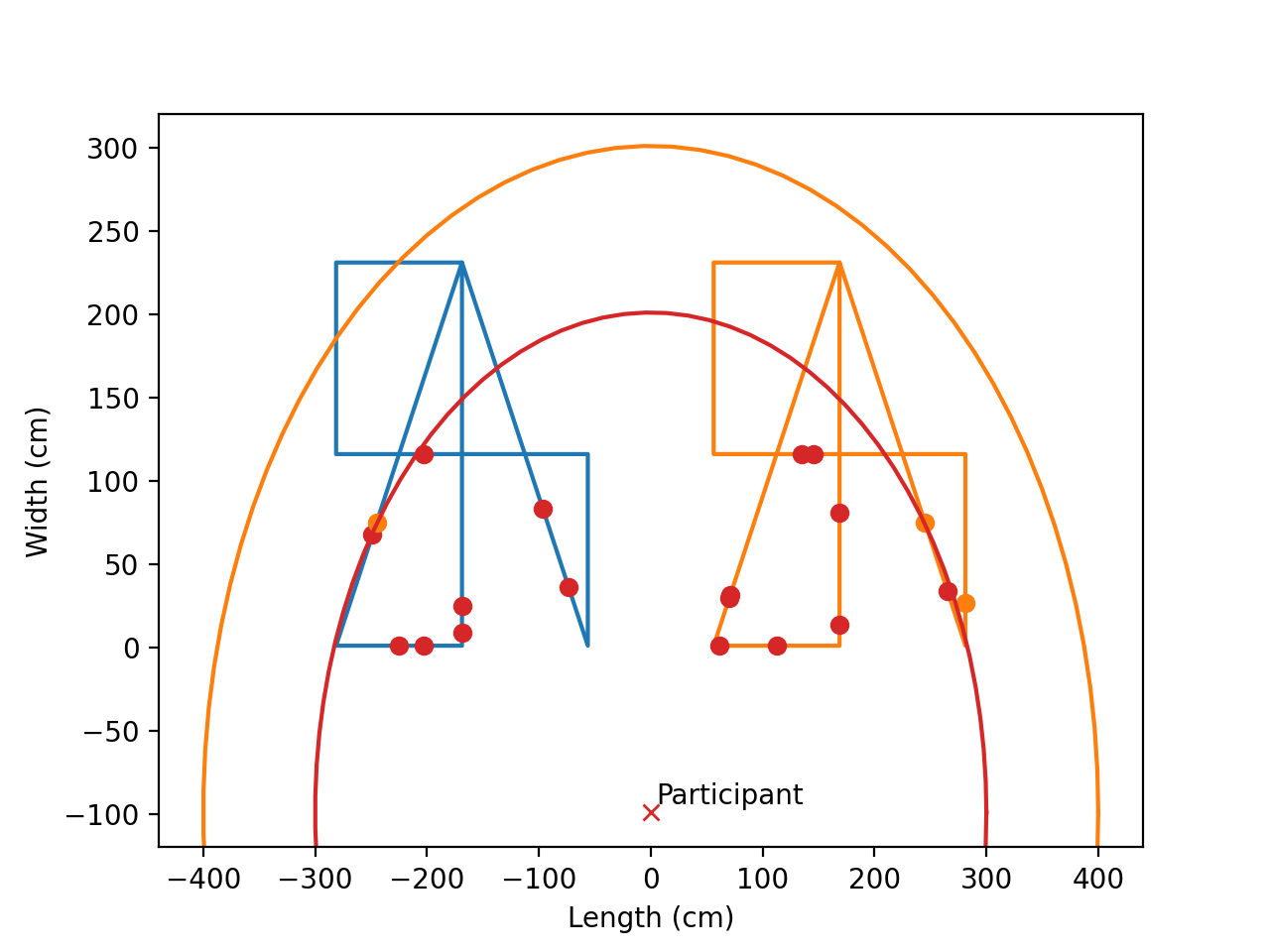}
        \caption{Setup B distance band region}
        \label{fig:swarm_rev_path_ranged_v1}
    \end{subfigure}
    \Description[Multiple drones experiment flight paths]{Multiple drones experiment flight paths.}
    \caption{Multiple drones experiment flight paths with discomfort button press positions in Red, the distance histogram showing button press distribution, and the banded distance regions.}
    \label{fig:swarm_path_experiments}
\end{figure*}

Figures \ref{fig:swarm_fwd_path_position_v1} and \ref{fig:swarm_rev_path_position_v1} show the result of the Experiment 2 Setup A and B multiple drone experiments respectively. The 2D flight path for the first and second UAV are plotted in Blue and Yellow respectively, with Blue arrows in the middle of each line segment indicating each UAV's direction of travel. The Red dots indicated the point during the flight when the participants pressed the discomfort button. For the multiple drones experiment setup, it was not possible to distinguish which of the UAVs caused the discomfort when the participants pressed the button since they only had one button to indicate discomfort. Therefore, the position of the two drones at each button pressed by the participants was plotted on each drone's trajectory. Similar to the observation in Experiment 1, there is a concentration of Red dots closer to the participant indicating increased discomfort when the UAVs are closer to the participants. Unlike in Experiment 1B, we did not observe any empathy effect for the observer.

Figures \ref{fig:swarm_fwd_button_press_histogram_v1} and \ref{fig:swarm_rev_button_press_histogram_v1} show the histogram distribution of the participants' button presses. The histogram plot started at 200 cm unlike in the first experiments because there were not enough button press points sub-200 due to the participants pressing the discomfort button much earlier indicating an earlier discomfort distance than in Experiment 1. Also, due to the 100 cm safety buffer between the participants and the drone flight path, there were no points below 100 cm. The histogram bins are at an interval of 100 cm, for easier quantisation. From the histogram plots, three zones emerged, the Red zone marking the point of most discomfort (<300 cm), the Yellow zone marking points of moderately acceptable discomfort (300 - 400 cm),  and the Green zone marking points of little or no discomfort (> 400 cm). Participants experienced discomfort earlier in the multiple drones test than in the single drone test suggesting that increasing the number of drones could increase co-located human discomfort. About 81 \% of the total button presses (31) occurred in the 200 - 300 cm zone and the remaining 19 \% was in the 300 - 400 cm zone. Figures \ref{fig:swarm_fwd_path_ranged_v1} and \ref{fig:swarm_rev_path_ranged_v1} show the Experiment 2A and 2B path plots respectively, each segmented by radial lines at distances of 300 cm in Red and 400 cm in Yellow colours with the button press points that fall within this being colour-coded to indicate which region they fall into.

\section{Discussion}
The result of our experiments suggested safe distances for a single drone should be 200 cm and for multiple drones should be 300 cm. These margins are higher than that observed by Acharya et al. \cite{Acharya2017}, 65.5 cm for the small UAV, and Wojciechowska et al. \cite{wojciechowska_collocated_2019}, UAV within personal space (50 to 121 cm) over intimate (< 50 cm) and social distances (> 121 cm). This difference may be because in the previous experiments, the UAVs were flown in autonomous mode whereas in our experiments they were flown manually by UAV pilots. This resulted in slightly different speeds of movement whereas in the previous studies, the UAV speed was fixed and predictable. Also, unlike the previous study where a single-path was used, we used a multiple-path approach. All of these could have contributed to making our participants uncomfortable earlier which resulted in the higher safe distance margin at 200 cm. We also observed in the multiple drones case that the safe distance margin was 300 cm. The increase in the number of drones caused the participants to become uncomfortable earlier. This suggests that UAV swarms require more safety margin or distance buffer than a single UAV in co-located human-UAV interaction. We could not find previous works to benchmark the multiple drones' distance margin against, and this may be one of the first works investigating multiple drone (or aerial swarm) proxemics to determine safe operating zones around co-located humans in order to minimise discomfort or anxiety.

We also observed in Experiment 1B the empathy phenomenon. We noticed that the participants of the Experiment Setup B experiment indicated discomfort when the UAV was close to them and when it was also close to the observer. Their sitting position meant they had a better view of the observer and they seemed to have reacted to drone proximity to the observer. This phenomenon was not observed in Experiment 1 Setup A probably because the observer (experimenter with camera) was outside the angle of view of the participants (Participant 1 position) being over 400 cm to the left-hand side of the participants as shown in Figure \ref{fig:layout}. Also, this phenomenon was not observed in the multiple UAV experiments.

The main limitations of this research included the lack of autonomous flight and the limited number of multiple UAV experiments conducted. The lack of autonomous UAV flights resulted in path navigation velocity being inconsistent due to manual pilot flight. Some path segments were slightly slower, and others were slightly faster as seen in Figure \ref{fig:forward_path_result_blindtest} and \ref{fig:reverse_path_result_blindtest}. Autonomy also has the added benefit of making the flight paths repeatable and at a fixed velocity. Unlike the single UAV experiments with six different experiment flight paths, in the multiple UAV experiments, only one experiment flight path was used thereby limiting the number of approach directions tested. Also, there was no blindfold test for the multiple UAV experiment.

Future works would address these limitations and investigate how other factors such as direction of movement, UAV visibility, noise levels, speed levels, and number of UAVs in a swarm affect co-located human discomfort and safety. We would also investigate the empathy effect further to determine if it is a real effect in human-swarm interaction. In addition to these, the current work developed a 2D model of safe zones, future work would investigate and develop a 3D model of safe zones for aerial swarms. Also, although specific safe zones were identified in our experiments, these may vary depending on several factors such as the application scenario, environment, type of platform, onboard autonomy, etc. Therefore, future work would aim to autonomously determine safe zones for the swarm to operate based on the application context.

\section{Conclusion}
This research investigated safe operating zones for single and multiple UAVs to guide the deployments of aerial swarms in co-located human environments. We conducted practical laboratory-based user studies (N=18) with real UAVs to determine when the co-located human was uncomfortable. The participants were asked to indicate their discomfort with the press of a button. The time at which they pressed the button was logged alongside the position of the drone at the time the button was processed. The results were analysed and presented. Our result showed that participants were more uncomfortable with multiple UAVs closer to them than with a single UAV. The safe operating zones for the single UAV was found to be a minimum of 200 cm and for multiple UAVs was a minimum of 300 cm. However, these may vary depending on several factors such as the application scenario, environment, type of platform, onboard autonomy, etc. Therefore, future work would aim to autonomously determine safe zones for multiple UAVs to operate based on the application context. The main limitations of this preliminary study included the lack of autonomous flight and the limited number of multiple UAV experiments conducted. Future works would address these, develop a 3D model of safe zones for aerial swarms, and also investigate other factors contributing to human discomfort with multiple UAVs such as noise, speed, visibility, and the number of UAVs.

\begin{acks}
The authors wish to thank Shantam Sridev for helping to fly the drone during the experiments. This project was supported by the EPSRC Smart Solutions Towards Cellular-Connected Unmanned Aerial Vehicles System (EP/W004364/1), the UKRI Trustworthy Autonomous Systems Hub (EP/V00784X/1), and Responsible AI UK (EP/Y009800/1).
\end{acks}

\bibliographystyle{ACM-Reference-Format}
\bibliography{ref.bib}

\end{document}